\documentclass[a4paper]{jpconf}
\usepackage[numbers]{natbib}
\usepackage{romannum}
\usepackage{amsmath,amssymb,amsfonts}
\usepackage{graphicx}
\usepackage{textcomp}
\usepackage{xcolor}
\usepackage{tikz}
\usetikzlibrary{shapes.geometric}
\usepackage{algorithm,algpseudocode}
\usepackage{url}
\usepackage{hyperref}
\usepackage{citesort}
\usepackage{pseudocode}
\begin{document}

\title{Reinforcement learning approach for multi-agent flexible scheduling problems}

\author{Hongjian Zhou$^1$, Boyang Gu$^2$ and Chenghao Jin$^3$}

\address{$^1$ University of Oxford, Oxford OX1 2JD, United Kingdom}
\address{$^2$ Imperial College London, London SW7 2BX, United Kingdom}

\ead{$^1$\href{mailto:hongjian.zhou@exeter.ox.ac.uk}{hongjian.zhou@exeter.ox.ac.uk}, $^2$\href{mailto:boyang.gu19@imperial.ac.uk}{boyang.gu19@imperial.ac.uk}, $^3$\href{mailto:steven_jin_gbhg@foxmail.com}{steven\_jin\_gbhg@foxmail.com}}

\begin{abstract}
Scheduling plays an important role in automated production. Its impact can be found in various fields such as the manufacturing industry, the service industry and the technology industry. A scheduling problem (NP-hard) is a task of finding a sequence of job assignments on a given set of machines with the goal of optimizing the objective defined. Methods such as Operation Research, Dispatching Rules, and Combinatorial Optimization have been applied to scheduling problems but no solution guarantees to find the optimal solution. The recent development of Reinforcement Learning has shown success in sequential decision-making problems. This research presents a Reinforcement Learning approach for scheduling problems. In particular, this study delivers an OpenAI gym environment with search-space reduction for JSSP and provides a heuristic-guided $Q$-Learning solution with state-of-the-art performance for Multi-agent Flexible Job Shop Problems. 
\end{abstract}
\begin{keywords}
scheduling, job shop, flexible job shop, reinforcement learning, heuristic-guided reinforcement learning, $Q$-learning, multi-agent system, optimization
\end{keywords}

\section{Introduction}
\label{section:intro}
Production scheduling is defined as an optimization of the allocation of resources over time among activities performed by machines. It is crucial to real-world production since the resources are always limited and require scheduling for optimal outcomes. In this article, we mainly discuss the Job Shop Scheduling Problem (JSSP), a very common application in production scheduling. 

Recently, many approaches are developed for JSSP. In order to study them, we propose an environment which can be used to test different algorithms and can be customized to deal with other variations of JSSP. We also present a new $Q$-learning algorithm which outperforms most traditional approaches.

The article is organized as follows: Section~\ref{section:related_work} gives a literature review of the related works of the scheduling problem, including the common models, related problems, and solutions. Section~\ref{section:problem_formulation} gives a review of the background knowledge to understand the model of JSSP. Section~\ref{section:JSSP_env} provides the detailed 
explanation of our designed environment for JSSP. Section~\ref{section:solution} explains our improved $Q$-learning algorithm. Section~\ref{section:evaluation} analyzes and evaluates the performance of our algorithm. Section~\ref{section:conclusion_and_future_work} concludes the whole article and discusses possible future works.

\section{Related Work}
\label{section:related_work}
One classical model of scheduling problems is the Markov Decision Process (MDP), since most scheduling problems make decisions more on the current state rather than on the entire history. Of course, mathematically most scheduling problems are not Markovian, so there will be some adjustments to the ordinary MDP model. Generalized semi-Markov Decision Process (GSMDP) is one of the suitable adjusted models \cite{messias2013gsmdps}, where it allows the enabled event to trigger at any time and optimize the makespan. When stochasticity is added to the model, GSMDP can use continuous phase-type distributions to simulate the duration, such as the duration time of the task \cite{younes2004solving}.

Before the introduction of reinforcement learning (RL) to the scheduling problem, many traditional approaches were developed. One of the famous scheduling problems is the Traveling Salesmen Problem (TSP), where Lin-Kernighan-Helsgaun TSP
solver was once the state-of-art heuristic for it \cite{helsgaun2017extension}. LKH solver transfers asymmetric problems into symmetric ones and uses the amount of violated constraints to scale the penalty. Another popular problem related to scheduling is the Resource-Constrained Project
Scheduling Problem (RCPSP). Kolisch extended the precedence based minimum slack priority rule (MSLK) to a precedence and resources based slack priority rule to replace RSM to solve RCPSP \cite{kolisch1996efficient}. If we relax the resource constraint and add stochasticity to the activities themselves, we get the Stochastic Scheduling Problem (SSP). There are many non-RL approaches to it. For example, the Rollout algorithm is used to solve the Quiz Problem, which is one variation of SSP \cite{bertsekas1999rollout}. There are also many heuristic approaches which describe different aspects of the stochastic features, which allows the usage of a computationally tractable closed-loop algorithm to solve SSP \cite{li2015solving}.

Another well-known area of scheduling problems is the Job Shop Scheduling Problem (JSSP), which we will further discuss it later in Section~\ref{section:JSSP_env}. {Milo{\v{s}}} introduced a classical model of it, including job-based representation, operation-based representation, and disjunctive graph-based representation \cite{vseda2007mathematical}, while actually there are many variations of JSSP. According to the number of operations, the number of available machines for each operation, and whether stochasticity is introduced to the problem, we have Parallel Machines Job Shop Scheduling Problem (JSSP-PM), Flexible Job Shop Scheduling (FJSSP), and Stochastic Flexible Job Shop Scheduling (SFJSSP) \cite{jimenez2012generic}. Wittrock introduced a non-RL algorithm to a variation of FJSSP, which decompose the problem into three sub-problems (machine allocation, order sequencing, and execution time) and each of these is solved using a fast heuristic \cite{wittrock1988adaptable}. There are also many RL-based algorithms developed for solving JSSP. Rinciog and Meyer developed the possibility of using RL-based algorithms to solve JSSP and gave a general design \cite{rinciog2022towards}. They separated MDPs for production scheduling into operation sequencing, routing before sequencing, interlaced routing and sequencing transport-centric routing and re-scheduling. They also classified RL algorithms along three discrete axes: value, policy or actor-critic methods; model-free or model-based methods; and on or off-policy methods. For more specific work, Cheng, Gen and Tsujimura introduced nine most common representations (job-based, operation-based, preference list-based, job pair relation-based, priority rule-based, disjunctive graph-based, completion time-based, machine-based, and random keys) of JSSP that are suitable for further Genetic algorithm (GA) work \cite{cheng1996tutorial}, and those representations are also suitable for other algorithms, both RL and non-RL ones. They also gave a full review of different variations of GA that can be applied to JSSP, including adapting problems to the genetic algorithms, adapting the genetic algorithms to problems, and adapting both the genetic algorithms and problems \cite{cheng1999tutorial}. Hasan, Sarker and Cornforth also gave a detailed improved GA algorithm which includes phenotypes creation, feasibility check, solution ranking or heuristic ordering, and generation updates (crossover and mutations) \cite{hasan2007hybrid}. Type-Aware Graph Attention (TAGAT) algorithm can also solve JSSP by using ScheduleNet, a decentralized decision-making policy that can coordinate multiple agents to complete tasks \cite{park2021schedulenet}. Besides those two algorithms, there are also many works on solving JSSP based on the $Q$-learning algorithm. The main variation of different $Q$-learning algorithms is they have different reward or cost strategies for different aims, such as makespan \cite{jimenez2012generic} or idle time \cite{reyna2015reinforcement}. Need to mention that besides JSSP, RL-based algorithms are also developed for other different scheduling problems. For example, Kuhnle et al. applied RL to create an adaptive production control system by the real-world example of order dispatching in a complex job shop \cite{kuhnle2021designing}.

\subsection{Flexible Job Shop Schedule Solutions}
There are three main approaches to solving the FJSSP: hierarchical approaches, integrated approaches, RL approaches \cite{jimenez2012generic}. 
\subsubsection{Hierarchical Approaches}
Hierarchical approaches decompose the problem into smaller problems with the intent of reducing complexity. The most common decomposition is to split the original question into assignments and sequencing parts. The assignments part determines the assignments of operations to machines. The problem is then transformed into a JSSP where each operation has an assigned machine for execution. Last, the sequencing part determines the sequence of execution. Two common algorithms adapting the hierarchical approaches are tabu search \cite{saidi2007flexible} and simulated annealing \cite{yazdani2009simulated}. 
Mohammad and Parviz \cite{saidi2007flexible} proposed a tabu search procedure to find the best sequence of operations and choice of machine alternatives simultaneously. The procedure in \cite{saidi2007flexible} uses a metaheuristic search method to find the best choice of machine's alternative for each job and operation sequence. These sequences are then improved iteratively by comparing them with neighbourhood solutions. The proposed algorithm finds success in solving smaller instances (less than 5 jobs and machines) of FJSSP but fails to solve larger instances (more than 5 jobs and machines) \cite{saidi2007flexible}. 
On the other hand, Yazdani and Gholami \cite{yazdani2009simulated} used a simulated annealing algorithm to address the FJSSP. They proposed a meta-heuristic algorithm to explore the solution space using a stochastic local search with probabilistic moves of avoiding local optima \cite{yazdani2009simulated}. Similar to \cite{saidi2007flexible}, the proposed algorithm finds success in solving smaller instances but struggles to find the optimal solution for larger instances. 
\subsubsection{Integrated Approaches}
Integrated approaches attempt to solve the problem as a whole instead of decomposing it. Common practices of integrated approaches are dispatching rules \cite{jun2019learning} and Genetic Algorithm \cite{pezzella2008genetic}. 
Jun and Lee proposed a Random Forest for Obtaining Rules for Scheduling (RANFORS) approach which consists of schedule generation, rule learning with data transformation and rule improvement with discretization \cite{jun2019learning}. The result shows success in capturing both explicit and implicit knowledge while outperforming prevalent dispatching rules in performance and robustness.
Alternatively, Pezzela and Morganti \cite{pezzella2008genetic} presented a genetic algorithm which integrates different strategies for generating the initial population and reproducing new ones. The results obtained are comparable to the best-known algorithms based on tabu search.
\subsubsection{RL Approaches}
RL approaches learn action-state values that give the expected utility of taking a given action in a given state, then choose the optimal pairs as the solution.
Both Reyna \cite{reyna_caceres_jimenez_reyes_2018} and Martinez \cite{martinez_nowe_suarez_bello_2011} used $Q$-learning to solve FJSSP. While Martinez decomposes the original FJSSP into assignment and sequencing, Reyna approaches FJSSP as one single problem. The results obtained from both approaches show success in smaller and medium-sized problems (less than 10 jobs and machines), outperforming classical methods such as genetic algorithm and tabu search. However, they struggle to find the optimal solution for large-sized problems (more than 10 jobs and machines).
In addition to $Q$-learning, researchers tried more advanced learning methods such as Deep Reinforcement Learning (DRL) to solve JSSP. DRL Tassel, Gebser \cite{tassel_gebser_schekotihin_2021} and Chang, He and Yu \cite{chang_yu_hu_he_yu_2022}  both used DRL to solve JSSP. While Tassel and Gebser focused on solving the classical JSSP problem where the machine and operation pairs are pre-determined, Chang, He and Yu focused on solving the dynamic FJSP with random job arrival. Tassel and Gebser used two Multi-Layer Perceptron (MLP) models for action selection and state-value prediction respectively. On the other hand, Chang, He and Yu used Deep $Q$-network (DQN) to learn the experience tuple with states, actions and rewards. Results obtained from both study show success in small, medium and large-sized problems and outperforming most classical approaches and RL approaches.

\section{Problem Formulation}
\label{section:problem_formulation}
Markov Decision Process(MDP) and its variations are the most common ways of describing scheduling problems, which we will introduce below. Note that all the stochastic process in this section is time-homogeneous.
\subsection{MDP}
The decision maker makes decisions as the system evolves through time. We denote $T$ as the time points where the state is observed and the action must be chosen, ie., the set of all decision epochs or stages. $T$ can either be discrete or continuous, with or without finite horizons. We will focus on the discrete case with an infinite horizon for the rest of the article. Conventionally, we let $T=\mathbb{N}$ in this case. For any $t\in T$, we denote $S_t$ the set of all possible state at time $t$ and $X_t$ the random variable of state at time $t$. For any $s\in S_t$, we denote $A_{s, t}$ the set of all applicable actions when state $s$ is observed at time $t$ and $A_t$ the random variable of action at time $t$. According to the action $a$ the agent chose, the system will transfer into another state in the next time point, ie., $s^{\prime}\in S_{t+1}$, with a probability $p(s, a, s^{'})$ and receive a reward. This stochastic process is a Markov Decision Process (MDP) if
\begin{equation}
\begin{aligned}
\mathbb{P}(X_{t+1}=s_{t+1} | X_0=s_0, A_0=a_0, \ldots, X_t=s_t, A_t=a_t)
=\mathbb{P}(X_{t+1}=s_{t+1} | X_t=s_t, A_t=a_t)\,,\\
\forall t,s,a\,.
\end{aligned}
\end{equation}
Here we denote 
\begin{equation}
    p(s_t, a_t, s_{t+1})=\mathbb{P}(X_{t+1}=s_{t+1} | X_t=s_t, A_t=a_t)\,,
\end{equation}
so clearly $\sum_{s_{t+1}\in S_{t+1}}p(s_t, a_t, s_{t+1})=1$.

To summarize, the standard discrete MDP consists of five components: Decision Epochs ($T$), State Space ($S$), Action Space ($A$), Transition Probability ($P$), and Reward Function ($R$) \cite{puterman1990markov}.
\subsection{Semi-MDP}
The main difference between MDP and semi-MDP is that we consider the time difference between the two decision epochs. We first introduce the Markov Renewal Process.

Suppose we have the same definition of $A_n, S, A$, we modified our definition of the set of decision epochs $T$ into the following: Instead of considering $T$ as consecutive time points, we consider only the time points where observation changes and action is made. For each $n\in\mathbb{N}$, we have a random variable $T_n\in [0,+\infty)$ s.t. $T_0\leq T_1\leq\cdots$.

The stochastic process $(X,T)$ is a Markov Renewal Process if
\begin{equation}
\begin{aligned}
\mathbb{P}(X_{n+1}=s_{n+1}, T_{n+1}-T_n\leq t | X_0=s_0, A_0=a_0, T_0=t_0, \cdots, X_n=s_n, A_n=a_n, T_n=t_n)\\
=\mathbb{P}(X_{n+1}=s_{n+1}, T_{n+1}-T_n\leq t | X_n=s_n, A_n=a_n)\,,\\
\forall n,s,a,t\,.
\end{aligned}
\end{equation}

If we define $(Y,B)=\{Y_t,B_t:t\geq 0\}$ as 
\begin{equation}
\begin{aligned}
Y_t = X_n \text{ if } T_n\leq t\leq T_{n+1}\,,\\
B_t = A_n \text{ if } T_n\leq t\leq T_{n+1}\,,
\end{aligned}     
\end{equation}
then it is considered a semi-MDP \cite{baykal2010semi}. It is called semi-MDP because the process is Markovian only at the specified jump instants. The advantage of semi-MDP compared to MDP is that we can simulate the stochastic duration time of an action by assigning the random variable $T_{n+1}-T_n$ a distribution.

\section{Job Shop Scheduling Problem Environment}
\label{section:JSSP_env}
In this section, we first introduce the basic structure of the JSSP and FJSSP environments. Then we describe our implementation of the FJSSP environment focusing on two areas: the structure of the environment and its interaction with the agent.
\subsection{JSSP and FJSSP}
Scheduling problems are a set of decision-making processes that aim to optimize objectives based on the allocation of resources. They are often characterized by the machine environment, the job characteristics and the objective functions \cite{herrmann_lee_snowdon_1992}. The general problem our environment implementation target is the Job Shop Scheduling Problems (JSSP), which involves a set of machines and jobs to minimize the makespan (the time it takes to finish all jobs). The constraints inherent in the definition of JSSP are the following \cite{jimenez2012generic}:
\begin{itemize}
    \item Operations from the same job can not be processed simultaneously.
    \item Machines can not process more than one operation each time.
    \item No interruption of operations is allowed.
    \item Operations from the same job must be processed in a predefined order.
    \item Machines are allowed to be idle.
    \item All jobs must be completed.
\end{itemize}

We formally define JSSP in the following mathematical formulation \cite{garey_johnson_1979}. Given a set of jobs $\{J_1, \dots, J_n\}$ and a set of machines $\{M_1, \dots, M_n\}$, each job $j_i$ consists of $m$ chained operations $\{O_{1Ji}, \dots, O_{nJi}\}$. The goal is to assign jobs to machines and find a schedule that minimizes the makespan $C_{max}$. The problem is proved to be NP-hard \cite{garey_johnson_sethi_1976}, and even state-of-the-art algorithms fail to consistently find optimal solutions \cite{jimenez2012generic}. 

\begin{figure}[htbp]
\centerline{\includegraphics[width=\textwidth]{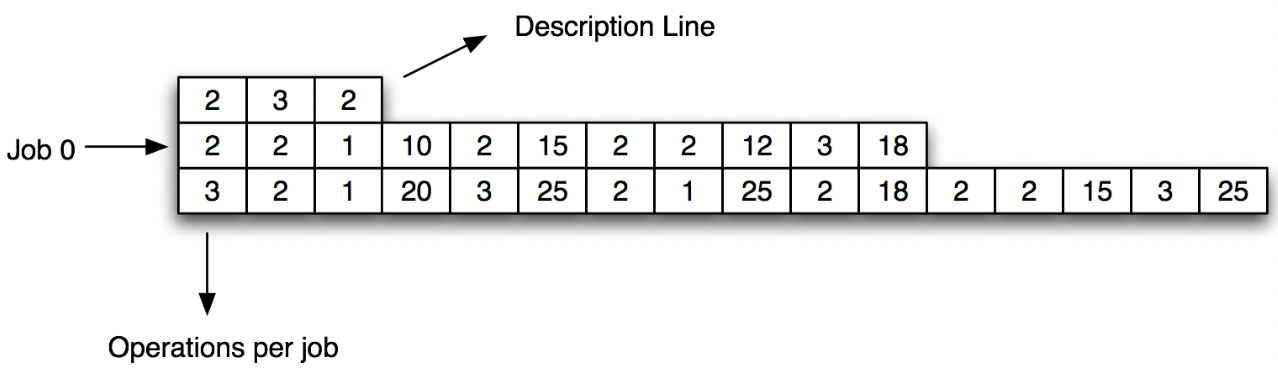}}
\caption{Flexible Job Shop Scheduling Problem Instance}
\label{fig:fjssp_instance}
\end{figure}

The specific scheduling problem that our RL solution solves is the Flexible Job Shop Scheduling Problem (FJSSP). As an extension of JSSP, FJSSP inherits the constraints of JSSP. 
In addition to the classical JSSP where job assignments are prescribed and machine-dependent, FJSSP allows assignments of operations to a set of candidate machines with different processing speeds \cite{rossi_boschi_2009}. Fig.~\ref{fig:fjssp_instance} is an example of an FJSSP instance \cite{beasley_1990}. In this example, multiple machines are capable of executing the same operation at different speeds, making this a more challenging scheduling problem than the classical JSSP.  

We consider FJSSP as a multi-agent problem, where multiple jobs are assigned to multiple machines at each timestep. The state-of-the-art solution for FJSSP is OR-Tools, which uses the CP-SAT Solver to find the optimal solution for FJSSP \cite{ortools}. 
In the next sections, we present an efficient JSSP environment for both RL and Deep RL with search-space reduction. 
\subsection{Environment Structure}
The environment structure consists of three components: a nested dictionary that stores the problem information, an array that represents the observation state and a system of dictionaries and arrays to reduce search space and represent the action state.
\subsubsection{Data Storage}
\begin{figure}[htbp]
\centering
\resizebox{0.7\textwidth}{!}{%
\begin{tikzpicture}[>=latex]
    \node (s0) at (0,10) [rectangle, draw=black, fill=white, align=left] {
    Instance1: \\ 
    2 3 \\
    2 2 1 10 2 15 2 2 12 3 18 \\
    3 2 1 20 3 25 2 1 25 2 18 2 2 15 3 25
    };
    \node (s1_1) at (0,0) [rectangle, draw=black, fill=white] {Job-Operation-Map};
    
    \node (s2_1) at (2,4) [rectangle, draw=black, fill=white] {Job$_1$};
    \node (s2_2) at (2,-4) [rectangle, draw=black, fill=white] {Job$_2$};
    
    \node (s3_1) at (5,6) [rectangle, draw=black, fill=white] {$\text{Operation}_1^1$};
    \node (s3_2) at (5,2) [rectangle, draw=black, fill=white] {$\text{Operation}_2^1$};
    \node (s3_3) at (5,-2) [rectangle, draw=black, fill=white] {$\text{Operation}_1^2$};
    \node (s3_4) at (5,-4) [rectangle, draw=black, fill=white] {$\text{Operation}_2^2$};
    \node (s3_5) at (5,-6) [rectangle, draw=black, fill=white] {$\text{Operation}_3^2$};
    
    \node (s4_1) at (8,7) [rectangle, draw=black, fill=white] {$\text{Machine}_1$};
    \node (s4_2) at (8,5) [rectangle, draw=black, fill=white] {$\text{Machine}_2$};
    \node (s4_3) at (8,3) [rectangle, draw=black, fill=white] {$\text{Machine}_2$};
    \node (s4_4) at (8,1) [rectangle, draw=black, fill=white] {$\text{Machine}_3$};
    \node (s4_5) at (8,0) [rectangle, draw=black, fill=white] {$\text{Machine}_1$};
    \node (s4_6) at (8,-2) [rectangle, draw=black, fill=white] {$\text{Machine}_3$};
    \node (s4_7) at (8,-3) [rectangle, draw=black, fill=white] {$\text{Machine}_1$};
    \node (s4_8) at (8,-5) [rectangle, draw=black, fill=white] {$\text{Machine}_2$};
    \node (s4_9) at (8,-6) [rectangle, draw=black, fill=white] {$\text{Machine}_2$};
    \node (s4_10) at (8,-8) [rectangle, draw=black, fill=white] {$\text{Machine}_3$};
    
    \node (s5_1) at (10,7) [rectangle, draw=black, fill=white] {$10$ $Ts$};
    \node (s5_2) at (10,5) [rectangle, draw=black, fill=white] {$15$ $Ts$};
    \node (s5_3) at (10,3) [rectangle, draw=black, fill=white] {$12$ $Ts$};
    \node (s5_4) at (10,1) [rectangle, draw=black, fill=white] {$18$ $Ts$};
    \node (s5_5) at (10,0) [rectangle, draw=black, fill=white] {$20$ $Ts$};
    \node (s5_6) at (10,-2) [rectangle, draw=black, fill=white] {$25$ $Ts$};
    \node (s5_7) at (10,-3) [rectangle, draw=black, fill=white] {$25$ $Ts$};
    \node (s5_8) at (10,-5) [rectangle, draw=black, fill=white] {$18$ $Ts$};
    \node (s5_9) at (10,-6) [rectangle, draw=black, fill=white] {$15$ $Ts$};
    \node (s5_10) at (10,-8) [rectangle, draw=black, fill=white] {$25$ $Ts$};
    
    \draw [->, ultra thick] (s1_1)--(s2_1);
    \draw [->, ultra thick] (s1_1)--(s2_2);
    
    \draw [->, very thick] (s2_1)--(s3_1);
    \draw [->, very thick] (s2_1)--(s3_2);
    \draw [->, very thick] (s2_2)--(s3_3);
    \draw [->, very thick] (s2_2)--(s3_4);
    \draw [->, very thick] (s2_2)--(s3_5);
    
    \draw [->, dashed] (s3_1)--(s4_1);
    \draw [->, dashed] (s3_1)--(s4_2);
    \draw [->, dashed] (s3_2)--(s4_3);
    \draw [->, dashed] (s3_2)--(s4_4);
    \draw [->, dashed] (s3_3)--(s4_5);
    \draw [->, dashed] (s3_3)--(s4_6);
    \draw [->, dashed] (s3_4)--(s4_7);
    \draw [->, dashed] (s3_4)--(s4_8);
    \draw [->, dashed] (s3_5)--(s4_9);
    \draw [->, dashed] (s3_5)--(s4_10);
    
    \draw [->, thick] (s4_1)--(s5_1);
    \draw [->, thick] (s4_2)--(s5_2);
    \draw [->, thick] (s4_3)--(s5_3);
    \draw [->, thick] (s4_4)--(s5_4);
    \draw [->, thick] (s4_5)--(s5_5);
    \draw [->, thick] (s4_6)--(s5_6);
    \draw [->, thick] (s4_7)--(s5_7);
    \draw [->, thick] (s4_8)--(s5_8);
    \draw [->, thick] (s4_9)--(s5_9);
    \draw [->, thick] (s4_10)--(s5_10);

    \draw [->,dashdotted] (s0)--node[fill=white]{initialize}(s1_1);
\end{tikzpicture}
}%
\caption{Data Structure Storing FJSSP Information}
\label{fig:job_operation_map}
\end{figure}
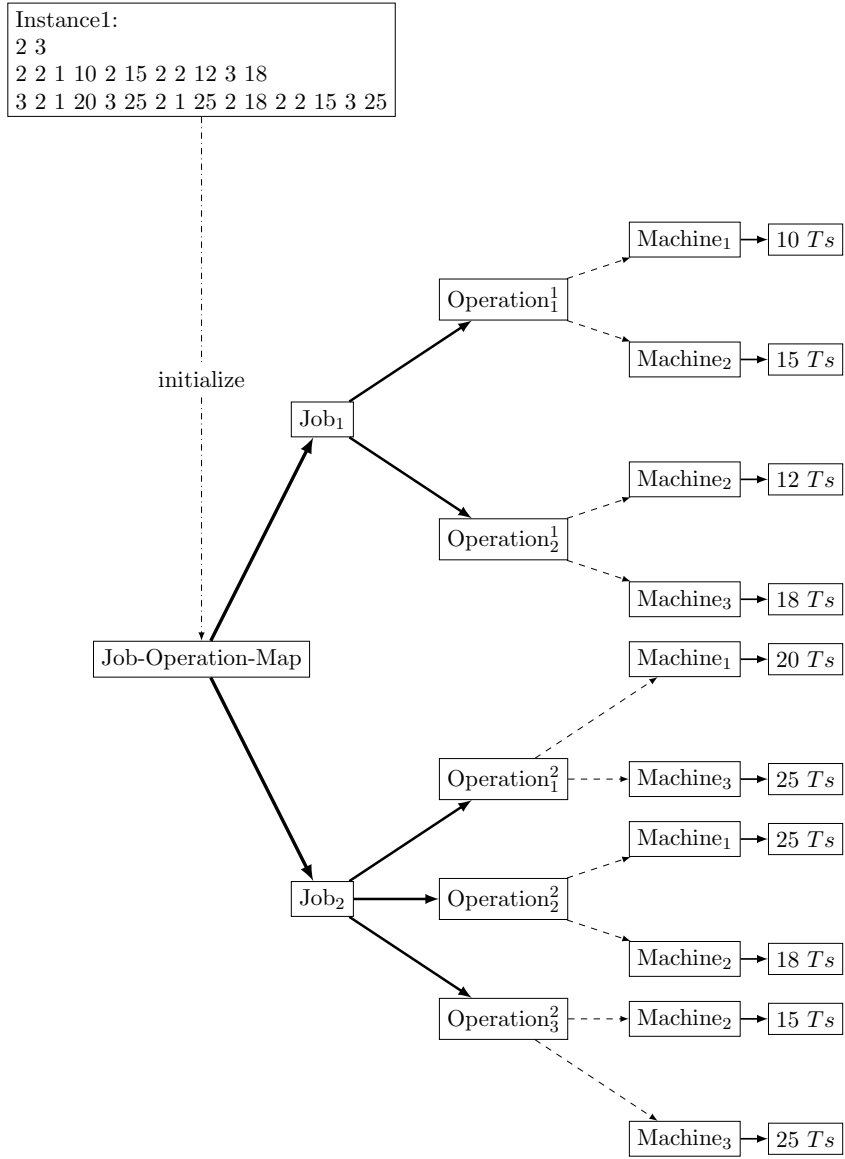

Our implementation of the FJSSP requires storing information such as problem formulation and constraints before initialization. In standard FJSSP solutions, the environment information is passed in as an instance file \cite{jimenez2012generic}. Fig.~\ref{fig:fjssp_instance} shows an example of an instance. The first line consists of three integers representing the number of jobs, the number of machines and the average number of machines per operation (optional) respectively. Each of the following lines has a distinct job description with the first integer being the number of operations and other integers being operation descriptions. For example, in Fig.~\ref{fig:fjssp_instance}, there are two jobs and three machines. The second line of the instance is a description of $J_1$. The first integer 2 of the second line indicates $J_1$ has two operations, $Op_{1J1}$ and $Op_{2J1}$. The second integer 2 indicates that two machines are capable of executing $Op_{1J1}$. The rest of the integers describe the machines and the times they take to execute the operations. For instance, the second line of the instance in full indicates that $J_1$ has two operations, $Op_{1J1}$ and $Op_{2J1}$. $Op_{1J1}$ can be carried out by two machines, $M_1$ and $M_2$, where they take 10 timesteps and 15 timesteps correspondingly. Additionally, there are two machines capable of executing $Op_{2J1}$: $M_2$ for 12 time steps and $M_3$ for 18 time steps. To initialize our environment, we store this information with a nested dictionary of arrays where job Ids and operation Ids are used as keys. Execution times are stored in arrays with machine Id being the index of the array shown in Fig.~\ref{fig:job_operation_map}. 
\subsubsection{Observation Space}
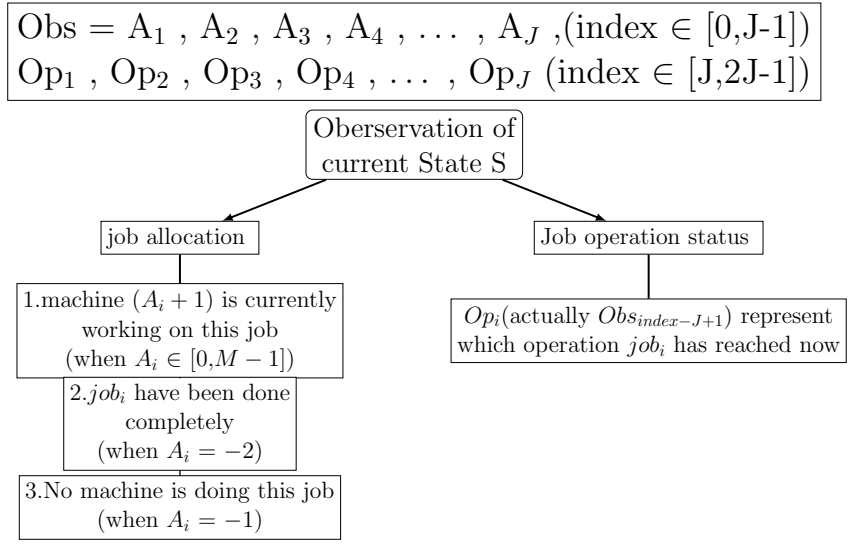
\begin{figure}[htbp]
\centering
\resizebox{0.7\textwidth}{!}{%
\begin{tikzpicture}[>=latex]
    \node (Allocation) at (0,2) [rectangle, draw=black, fill=white, font=\large, scale=1.5,align=center]{
       Obs = A$_1$ , A$_2$ , A$_3$ , A$_4$ , $\dots$ , A$_J$ ,(index $\in$ [0,J-1])\\
        Op$_1$ , Op$_2$ , Op$_3$ , Op$_4$ , $\dots$ , Op$_J$ (index $\in$ [J,2J-1])
    };
    
    \node (s1) at (0,0) [rectangle,rounded corners,text centered,draw=black,fill=white,font=\large,scale=1.2,align=center] {
        Oberservation of \\
        current State S
    };
    
    \node (s2) at (-5,-2) [rectangle,text centered,draw=black,fill=white,font=\large,scale=1,align=center] {
        job allocation
    };
    \node (s3) at (5,-2) [rectangle,text centered, draw=black, fill=white, font=\large,scale=1, align=center] {
        Job operation status
    };
    \node (s2_1) at (-5,-4)  [rectangle,text centered,draw=black,fill=white,font=\large,scale=1,align=center]{
        1.machine ($A_i+1$) is currently\\
        working on this job\\
        (when $A_i$ $\in$ [$0$,$M-1$])
    };
    \node (s2_2) at (-5,-6)  [rectangle,text centered,draw=black,fill=white,font=\large,scale=1,align=center]{
        2.$job_i$ have been done \\
        completely\\
        (when $A_i$ $=$ $-2$)
    };
    \node (s2_3) at (-5,-7.8)  [rectangle,text centered,draw=black,fill=white,font=\large,scale=1,align=center]{
        3.No machine is doing this job \\
        (when $A_i$ $=$ $-1$)
    };
    \node (s3_1) at (5,-4) [rectangle,text centered,draw=black,fill=white,font=\large,scale=1,align=center]{
        $Op_i$(actually $Obs_{index-J+1}$) represent\\
        which operation $job_i$ has reached now
    };
    
    \draw [->,very thick] (s1)--(s2);
    \draw [very thick] (s2)--(s2_1);
    \draw [thick] (s2_1)--(s2_2);
    \draw [thick] (s2_2)--(s2_3);
    
    \draw [->,very thick] (s1)--(s3);
    \draw [very thick] (s3)--(s3_1);
\end{tikzpicture}
}%
\caption{Observation Space of FJSSP}
\label{fig:observation_space}
\end{figure}

The observation space of the FJSSP environment is defined as the set of all possible observations by an agent at any given state. The implementation follows the Box Space type required for OpenAI gym environments \cite{brockman2016openai}. A typical observation of an FJSSP environment is two arrays with length $n$, where $n$ is the number of jobs declared by the problem formulation. With job Id being the index, each array describes the allocation status and the operation status of jobs accordingly Fig.~\ref{fig:observation_space}. For example, one of the observations of the environment in Fig.~\ref{fig:fjssp_instance} is $((1, 0),(0, 1))$, indicating that $M_2$ is working on $Op_{1J1}$ of $J_1$ and $M_1$ is working on $Op_{2J2}$ of $J_2$. Note that our true implementation merges two arrays into one since the immutability of the lengths of arrays guarantees that no ambiguity is introduced by the merge.
\subsubsection{Action Space}
\begin{figure}[htbp]
\centering
\resizebox{0.7\textwidth}{!}{%
\begin{tikzpicture}[>=latex]
     \node (s0) at (0,0) [rectangle, draw=black, fill=white,font=\large,scale=1.5,minimum width=2.4cm]{
        Allocation$_1$
     };
     \node (s1) at (0,-0.893) [rectangle, draw=black, fill=white,font=\large,scale=1.5,minimum width=2.4cm]{
        Allocation$_2$
     };
     \node (s2) at (0,-1.786) [rectangle, draw=black, fill=white,font=\large,scale=1.5,minimum width=2.4cm]{
        Allocation$_3$
     };
     \node (s3) at (0,-2.679) [rectangle, draw=black, fill=white,font=\large,scale=1.5,minimum width=2.4cm]{
        Allocation$_4$
     };
     \node (s4) at (0,-3.572) [rectangle, draw=black, fill=white,font=\large,scale=1.5,minimum width=2.4cm,minimum height=0.6cm]{
         ... 
     };
     \node (s5) at (0,-4.465) [rectangle, draw=black, fill=white,font=\large,scale=1.5,minimum width=2.4cm]{
        Allocation$_N$
     };
     
     \node (sgg) at (7,3) [rectangle, draw=black, fill=white, font=\large,scale=1.5,align=center]{
        if action(an index)$=$i ,\\ 
        then take action in the $i^{th}$ allocation.
     };
     
     \node (s6_1) at (6.66,1.3) [rectangle, draw=black, fill=white, font=\large, scale=1.5]{
            $\omega_i^1$ , $\omega_i^2$ , $\omega_i^3$ , $\omega_i^4$ , ... , $\omega_i^J$
     };
     \node (s0_1) at (6.26,0) [rectangle, draw=black, fill=white, font=\large, scale=1.5]{
            $3$ , $2$ , $0$ , $-1$ , ... , $1$ 
     };
     \node (s5_1) at (7,-4.465) [rectangle, draw=black, fill=white, font=\large, scale=1.5]{
            $\omega_N^1$ , $\omega_N^2$ , $\omega_N^3$ , $\omega_N^4$ , ... , $\omega_N^J$
     };
     \node (s_ex) at (6.62,-2.3) [rectangle, draw=black, fill=white, font=\large, scale=1.5, align=left]{
            $\omega_i^j$ represent in allocation $i$ ,\\
            job $j$ is :
     };
     \node (s_ex2) at (6.62,-7.2) [rectangle, draw=black, fill=white, font=\large, scale=1.5, align=left]{
        1.allocated to machine ($\omega_i^j+1$) (when $\omega_i^j$ $\in$ [$0$,$M-1$])\\
        2.is currently waiting for available machine ($\omega_i^j$ $=$ $-1$)
     };
     
     \draw [->,very thick] (s_ex)--(11.8,-2.3)--(11.8,-5.85);
     \draw [dashed] (s0)--(s0_1);
     \draw [dashed] (s5)--(s5_1);
     
\end{tikzpicture}
}%
\caption{Legal Allocations}
\label{fig:legal_allocations}
\end{figure}
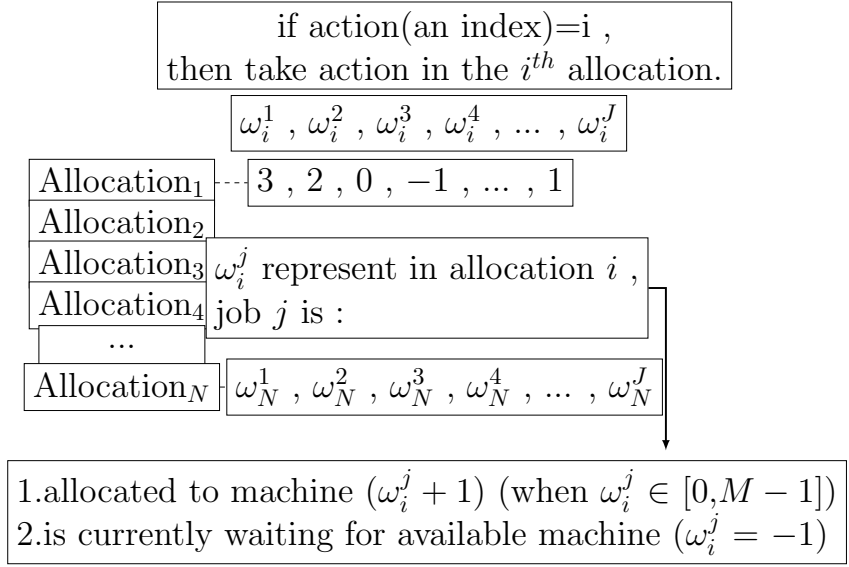

The action space of the FJSSP environment is defined as the set of all possible actions by an agent at each state. To achieve better performance, we reduce our action space to exclude all unnecessary actions. As a result of such reduction, the action space is dependent on the state of the environment. Furthermore, we generate a new set of legal allocations Fig.~\ref{fig:legal_allocations} when the environment enters a new state. Legal allocation is defined as a set of executable and reasonable assignments of jobs to machines given the current state of the environment. For example, one of the legal allocations for the environment in Fig.~\ref{fig:fjssp_instance} at state $S_0$ (timestep 0) is $[1, 0]$ which represents the assignments of $J_1$ to $M_2$ and $J_2$ to $M_1$. We store the legal allocations in an array and define the action to be the index of the allocation in the legal allocation array. Subsequently, the action space follows the Discrete Space type of OpenAI gym. The range of our action space is the length of the legal allocation array at each state.
\subsection{Environment Interaction}
The environment interaction consists of three components: a method of allocation generation and filtering to reduce search space, a step function with observation reduction to interact with the agent and a reward function to evaluate decisions made by the agent.
\subsubsection{Search space Reduction}
The action space can become extensively large if no filtering and restrictions are applied. Subsequently, we need to reduce the search space of  our agent to achieve better results in shorter times. We implement our search space reduction method as the generation of legal allocation at each state after an action is performed by an agent. Legal allocation is defined as a set of executable and reasonable assignments of jobs to machines given the current state(see \eqref{eq:allocation}). $A_k=j$ indicates the assignment of $J_k$ to $M_j$ or a wait action when $j=-1$. The search space (action space) of our agent at any given state is limited to the legal allocations generated. An allocation is legal if and only if the allocation is executable and reasonable. 
\begin{gather}
\label{eq:allocation}
    \left[A_1, A_2\dots, A_n\right]\\
    A_k \in \{1, \dots ,m\}\cup\{-1\}, k \in \{1, \dots ,n\}\nonumber\\
    n = \text{number of jobs}, m = \text{number of machines}\nonumber
\end{gather}

We say an allocation is executable if it satisfies all four of the following conditions for any $k \in \{1, \dots ,n\}$:
\begin{enumerate}
    \item [1.] If $A_k \neq -1$ The ongoing operation of $J_k$ can be executed by $M_{A_k}$ at current state of the environment.
    \item [2.] $J_k$ is not assigned or finished at the current state.
    \item [3.] $M_{A_k}$ is not assigned at the current state.
    \item [4.] No duplicate assignments of jobs or machines are present in the allocation.
\end{enumerate}

We say an allocation is reasonable if it satisfies:
\begin{enumerate}
    \item [1.] $A_k \neq -1 \forall k \in \{1, \dots ,n\} $ when all jobs or machines are idle.
\end{enumerate}
\subsubsection{Step Function}
The step function is mainly used by the agent to interact with the environment. Upon calling the step function, the agent executes an action and receives the observation of the next state, the reward of the executed action and a Boolean representing if all jobs are finished.  The step function consists of two processes: environment update and observation space reduction.

The environment update process first converts the action to a legal allocation and then assigns jobs to target machines accordingly.

The observation space reduction method aims to decrease the number of observations by eliminating intermediate states between valid states. We define valid states as states that contain any legal allocation other than wait. For example, if all jobs or machines are occupied after an action is executed, all proceeding states will be skipped until one of the jobs or machines become available. This process is implemented to accelerate the RL solution. 
\begin{figure}[htbp]
\centering
\resizebox{0.7\textwidth}{!}{%
\begin{tikzpicture}[>=latex]
    \node (s1) at (0,0) [rectangle,rounded corners,text centered,draw=black,fill=white,font=\large,scale=1] {State S};
    \node (s2) at (0,-1.3) [trapezium, trapezium left angle = 70,trapezium right angle=110,text centered,draw=black,fill=white,font=\large,scale=1] {Input};
    \node (s3) at (0,-3.3) [rectangle,text centered,draw=black,fill=white,font=\large,scale=1] {Time step + 1 and Update the state now};
    \node (s4) at (0,-5.3) [diamond,draw=black,fill=white,font=\large,scale=1] {
        Check 
    };
    \node (s5) at (0,-7.3) [rectangle,text centered,draw=black,fill=white,font=\large,scale=1] {Generate legal action list};
    \node (s6) at (0,-10) [diamond,draw=black,fill=white,font=\large,scale=1,align=center] {
        Legal action \\
        existing ?
    };
    
    \node (s7) at (0,-13.3) [rectangle,rounded corners,text centered,draw=black,fill=white,font=\large,scale=1] {new state S'};
    
    \draw [->,very thick] (s1)--(s2);
    \draw [->,very thick] (s2)--(s3);
    \draw [->,very thick] (s3)--(s4);
    \draw [->,very thick] (s4)--(s5);
    \draw [->,very thick] (s5)--(s6);
    \draw [->,very thick] (s6)--node[right,font=\large,scale=1]{Yes}(s7);
    \draw [->,very thick] (s4)--(-4.7,-5.3)--node[fill=white,font=\large,scale=1]{
        If all work have finished
    }(-4.7,-13.3)--(s7);
    \draw [->,very thick] (s6)--(5.7,-10)--node[fill=white,font=\large,scale=1]{
        No
    }(5.7,-3.3)--(s3);
    
    \node at (0,-14) [font=\large,scale=1]{
        Return Observation, Reward, list of Jobs had done
    };
    
\end{tikzpicture}
}%
\caption{Step Function}
\label{fig:step_function}
\end{figure}
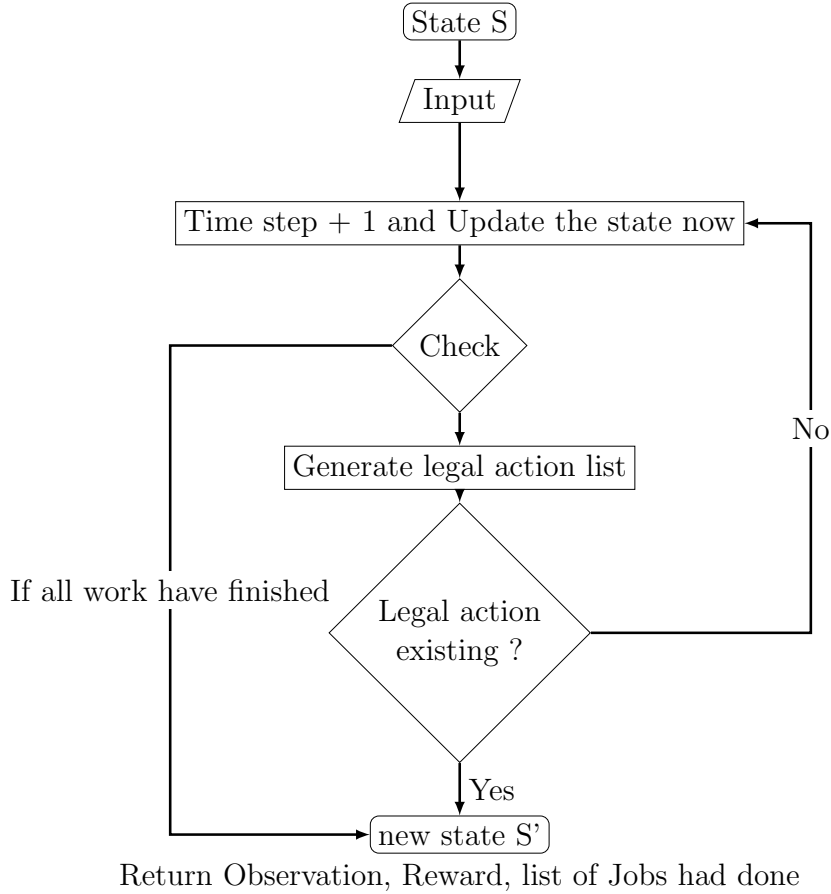
\subsubsection{Reward}
The metric we use to evaluate solutions is the maximum makespan across all jobs (the time it takes to finish all jobs). As a result, we design our reward function as
\begin{equation}
\label{eq:reward}
    R_{(s,a)} \gets T_{s} - T_{s'}\,,
\end{equation}
where $s^{\prime}$ is next state of state $s$ after execution of action $a$.

As a result of our evaluation metric (maximum makespan), the reward function is sparse and only provides feedback when all jobs are finished. The rewards received do not translate to the impact of the actions until the end of each episode. Instead, under convergence assumption, these rewards will be used in RL solutions and converge to measurements of impact. For example, using $Q$-Learning, the $Q$-value of a state action pair will converge to the makespan between the current state and the end state. Hence why convergence time is an important attribute in our RL solutions.

\section{Solution}
\label{section:solution}
Our implementation of the RL solution consists of three components: An $Q$-learning algorithm that learns the $Q$-value of state action pairs, a backward pass method that generates heuristic values to accelerate $Q$-learning and an instance division method to reduce environment space.
\subsection{Reinforcement Learning and $Q$-Learning}
One of the classical method to solve MDP-based problem is to use the $Q$-learning algorithm. The $Q$-value is defined as the expected discounted reward of the action at current state. In other words,
\begin{equation}
\label{eq:q_value}
    Q(s,a)=R_{(s,a)}+\gamma\cdot \max_{a^{'}}{Q(s_{next}, a^{'})}\,,
\end{equation}
where $\gamma$ is the discounted factor and $r$ is the reward of the action. In our case, the reward is the negative of the time from now to the nearest time that can do actions other than wait as discussed in the previous section. We are actually approximating the negative of the minimum makespan by the $Q$-value, so we set the discounted factor $\gamma = 1$.

In the $Q$-learning algorithm, we use $\epsilon$-greedy policy to choose the action and a updating rule to compute the $Q$-values. Instead of \eqref{eq:q_value}, every time we observe the state $s$ and take action $a$, we find the next state $s^{'}$ and update the $Q$-value $Q(s,a)$ by
\begin{equation}
\label{eq:q_learning}
    Q(s,a)\gets Q(s,a)+\alpha(R_{(s,a)}+\gamma\cdot\max_{a'}Q(s',a')-Q(s,a))\,,
\end{equation}
where $R_{(s,a)}+\gamma\cdot\max_{a'}Q(s',a')$ represents the expected reward of the action $a$ and we use it to update the $Q$-value but with a learning rate $\alpha$. It is proved that this updating rule will lead to a convergence to the actual $Q$-values \cite{watkins1992q}.
\subsection{Heuristic-Guided Q-Learning}
Based on the update rule of $Q$-learning, the algorithm converges when the reward of the current state action pair equals the difference between the $Q$-value of the current state action pair and the maximum $Q$-value of the next state. Under the convergence assumption, the $Q$-value of a state action pair is approximately the makespan between that state and the end state.

One of the challenges of our RL solution is the large convergence time. In spite of the space reduction methods, the issue of large search space is inherent to the complexity of large environments. Subsequently, to accelerate our solution for large environments, we implement a backward pass method to prepopulate the $Q$-table with heuristic values (Algorithm \ref{alg:prepopulate_q}). At the end of each episode, the prepopulate algorithm passes the cumulative rewards backward. Since rewards are defined as the timestep difference between the current and next state, the cumulative reward at any given state equals the makespan between that state and the end state with the actions taken in the current episode. Upon receiving the cumulative rewards, the algorithm updates the $Q$-value of the state action pair if the received value is more optimal. When the algorithm finishes, all visited state action pairs will hold a heuristic value equal to the most optimal makespan between that state and the end state among all episodes. There are two advantages of using the backward pass method. 
First, by prepopulating the $Q$-table with estimations, we can set a lower bound for the estimate of optimal future value at each state and hence avoid exploring sub-optimal search space. 
Second, the action space becomes extensively large for many FJSSP. Thus, it is hard to translate the impact of actions through the bellman equation update. Using backward pass, we can translate the impact of good actions immediately as they lead to a shorter makespan, hence a more optimal heuristic. 
\begin{algorithm}
\caption{Prepopulate $Q$-table}
\label{alg:prepopulate_q}
\begin{algorithmic}[1]
\State $e \gets \textbf{number of episodes}$
\While{$e \neq 0$}
\State $[\;(s_1,a_1), \dots, (s_n,a_n)] \gets \text{state action pairs visited}$
\State $[\;(r_1), \dots, (r_n)] \gets \text{all rewards received}$
\State $r_c \gets 0$ \Comment{cumulative reward}
\For{$k \gets n$ to $1$}
\If{$Q[(s_k, a_k)] \le r_c$}
    \State $Q[(s_k, a_k)] \gets r_c$
\EndIf
\State$r_c \gets r_c + r_k $
\EndFor
\EndWhile
\end{algorithmic}
\end{algorithm}
\subsection{Instance Division}
Prepopulation algorithm offers a way to decrease the convergence time of $Q$-learning. However, for extreme large instances, even prepopulation algorithm will fail to converge in a acceptable amount of time, so we introduce another algorithm called Instance Division to decrease the complexity of the instance. The idea is to split the instance into $n$ different sub-instances and denote them as instance$_1$, instance$_2$, ..., instance$_n$, then we compute instance$_1$ and get the best policy, and extend instance$_1$ to come up with a new instance(called instance$_{1,2}$) containing instance$_1$ and instance$_2$ and use the best policy from instance$_1$ as the fixed first half policy and only consider new actions when information of the second sub-instance comes. We keep doing so until we integrated all sub-instances.

We first need to create the sub-instances from the original instance. In the JSSP case, it can be done via the number of operations or the expected duration of operations. If we split the instance by the number of operations, for each job, each sub-instance will have the same number of operations if possible; if we split it by the expected duration time for each operations, we first calculate the expected duration time and it can be done in many different way, such as mean operation time or the maximum operation time of all available machines for this operation, then we will split the instance where we make sure that all sub-instances have the same expected duration time for all jobs if possible. For example, if we split the instance from Fig.~\ref{fig:fjssp_instance} into $2$ sub-instances, if it is done by the number of operations, the first sub-instance(instance$_1$) will contain only information of the first operation of the first job and the first and second operations of the second job, and the second sub-instance(instance$_2$) will contain all information of the second operation of the first job and the third operation of the second job; if it is done by the mean duration time of operations, we will calculate the mean duration time of each operations first. Let $t_{i,j}$ denote the mean duration of $j$-th operation of $i$-th job, in this case we have $t_{1,1}=12.5$, $t_{1,2}=15$, $t_{2,1}=22.5$, $t_{2,2}=21.5$, and $t_{2,3}=20$. We then set instance$_1$ contains all information of the first job and only the first and the second operations of the second job, and instance$_2$ contains only the third operation of the second job, so instance$_1$ has first job expected duration time $12.5+15=27.5$ and second job expected duration time $22.5+21.5=44$, and instance$_2$ has second job expected duration time $20$, which is the most even separation by the mean operation time.

When the sub-instances are created, we need to extend the best policy of the instance$_1$ into the instance$_{1,2}$ and so on until we get a best policy for instance$_{1,k}$, which is shown in Algorithm \ref{alg:instance_division}. Note that instance$_{1,k}$ represents the combined instance of all instances from instance$_1$ to instance$_k$. The key point of it is the get\_best\_policy function in Algorithm \ref{alg:get_best_policy}. The idea of get\_best\_policy function is that every time we choose an action, we need to make sure that this action coincide with the previous best policy from the previous instance. For example, suppose we have the best policy of instance$_1$ as policy$_1$, the new best policy of instance$_{1,2}$ must contain the actions in policy$_1$. The updating part of get\_best\_policy function(line $4$ to $7$ of Algorithm \ref{alg:get_best_policy}) depends on policy choosing algorithm, such as ordinary $Q$-learning or prepopulate $Q$-table.
\begin{algorithm}
\caption{Instance Division}
\label{alg:instance_division}
\begin{algorithmic}[1]
\State $[\text{instance}_1, \text{instance}_2, \cdots, \text{instance}_n]$ $\gets$ sub-instances
\State policy $\gets$ the best policy of instance$_1$
\For{$k \gets 2$ to $n$}
\State previous policy $\gets$ policy
\State current instance $\gets$ instance$_{1,k}$
\State policy $\gets$ get\_best\_policy(current instance, previous policy)
\EndFor
\end{algorithmic}
\end{algorithm}
\begin{algorithm}
\caption{get\_best\_policy(instance, previous policy)}
\label{alg:get_best_policy}
\begin{algorithmic}[1]
\State $e \gets \textbf{number of episodes}$
\For{k $\gets$ $1$ to $e$}
\State reset the state into the initial state
\While{this episode hasn't finished}
\State action $\gets$ action chosen from the policy where the previous policy is satisfied
\State update the state and the policy
\EndWhile
\EndFor
\State return the policy
\end{algorithmic}
\end{algorithm}

\section{Evaluation}
\label{section:evaluation}
In this section, we first describe the benchmark instances of FJSSP we use to evaluate our solution. Then we outline the baseline algorithms used for comparison. Last, we analyze the performance of our solution in comparison with other solutions under different environments. 
\subsection{Benchmark Instances}
To evaluate our RL solution against baseline solutions, we choose four distinct instances from OR-Library \cite{beasley_1990}. Instances are chosen to represent different levels of complexity. As a result of the instance selection procedure, four environments are created for evaluation:
\begin{enumerate}
    \item [1.] $Env_{s}$ with $3$ machines, $2$ jobs and a maximum operation count of $3$.
    \item [2.] $Env_{m}$ with $6$ machines, $6$ jobs and a maximum operation count of $6$.
    \item [3.] $Env_{l}$ with $6$ machines, $10$ jobs and a maximum operation count of $6$.
    \item [4.] $Env_{xl}$ with $11$ machines, $10$ jobs and a maximum operation count of $10$.
\end{enumerate}

As the number of jobs and machines increases, the complexity of FJSSP increases. These instances are used to evaluate the accuracy (ability to find optimal schedule) and efficiency (convergence time) of our solution.

To evaluate our RL solution against other RL literature, we used three sets of instances:
\begin{enumerate}
    \item [1.] Brandimarte instances \cite{brandimarte_1993}, Mk04 - Mk05, used in \cite{martinez_nowe_suarez_bello_2011}.
    \item [2.] Hurink instances \cite{hurink_jurisch_thole_1994}, la01 - la05, used in \cite{reyna2015reinforcement}.
    \item [3.] Taillard's instances \cite{taillard_1993}, Ta01 and Ta41, used in \cite{tassel_gebser_schekotihin_2021} and \cite{reyna_caceres_jimenez_reyes_2018}.
\end{enumerate}
\subsection{Baseline Selection}
To analyze the performance of our RL solution, we compared it against five different baseline solutions. These solutions include basic solutions such as Random Sampling, First In First Out (FIFO), Most Work Remaining (MWKR) and state-of-the-art solutions such as Genetic Algorithm and OR-Tools.
\subsubsection{Random Sampling (RS)}
Random Sampling (RS) is the basic algorithm for trying different schedules and choosing the optimal one. At each state of the environment, random action is sampled from the action space until all jobs are finished. The design of our environment guarantees the action chosen is executable and reasonable. Hence the random sampling algorithm will always reach the state when all jobs are finished. This process is repeated for multiple episodes and when all episodes finish, the best set of actions (schedule with the minimum makespan) among all episodes is chosen as the solution of the algorithm.
\subsubsection{FIFO and MWKR}
Two of the popular baseline algorithms for JSSP are Most Work Remaining (MWKR) and First In First Out (FIFO) \cite{tassel_gebser_schekotihin_2021}.
In MWKR, jobs with more left-over processing time left have higher priority. Using the observation of each state, the left-over time of each job can be calculated by taking the sum of time duration from the current operation to the last operation. For each state of the environment, jobs are assigned in descending order of the left-over time.
In FIFO, jobs with more waiting time have higher priority. The waiting time can be retrieved by taking the difference between the current timestep and the timestep the job was last assigned. For each state of the environment, jobs are assigned in descending order of the waiting time.
\subsubsection{Genetic Algorithm (GA)}
The genetic algorithm is a metaheuristic inspired by the process of natural selection. When used to solve JSSP, it contains the following steps:
\begin{enumerate}
    \item [1.] Encode some random scheduling as the first generation.
    \item [2.] Apply crossover or mutation on the previous generation to get the new generation.
    \item [3.] Repeat step $2$ until no great improvements are made.
    \item [4.] Decode the last generation to get the optimal scheduling.
\end{enumerate}
Note that the encoding step relies on the representation of the scheduling so there are several ways to do it \cite{cheng1996tutorial}.
\subsubsection{OR-Tools}
The OR-Tools is a state-of-art solution for solving complex problems such as vehicle routing, linear programming, and Scheduling \cite{ortools}. For each instance, our implementation of the solution for FJSSP with OR-Tools consists of the following steps:
\begin{enumerate}
    \item [1.] Model FJSSP with information from instance file.
    \item [2.] Create machine constraints.
    \item [3.] Initialize a makespan objective.
    \item [4.] Solve the model with CP-SAT solver.
\end{enumerate}
\subsection{Heuristic Guided and Instance Division}
This section examines the impact of our two improvement methods, heuristic-guided $Q$-learning with backward pass and instance division. We present the algorithm's performances with and without the improvement methods. 
\subsubsection{Heuristic Guided}
To shorten the convergence time, we implemented a heuristic-guided $Q$-learning method which uses a backward pass to prepopulate the $Q$-value table with estimations of the $Q$-value. Our reward function is sparse within episodes which causes the impact of the agent's decisions to not fully translate. As a result, the convergence time increases exponentially when the number of jobs and machines increases. To help the algorithm converge faster, we fill the $Q$-value table with estimations of the state-action pair: the makespan between the current state and the end state. Our experiments, Table~\ref{tab:Convtimes}, show that this approach efficiently shortens the convergence time for small and medium-sized problems. On top of that, classical $Q$-Learning is not able to converge while solving large and extra large-sized problems yet heuristic-guided $Q$-Learning successfully converges to the optimal solution in an efficient time. 
\begin{table}[htbp]
\caption{Convergence Time (s)}
\label{tab:Convtimes}
\begin{center}
\begin{tabular}{lll}
\br
Env &
Classical $Q$-Learning &
Heuristic Guided $Q$-Learning
\\
\mr
$Env_{s}$&  5& 0 \\
$Env_{m}$&  133& 5 \\
$Env_{l}$&  NA& 127 \\
$Env_{xl}$&  NA& 83 \\
\br
\multicolumn{3}{l}{All results rounded to the closest integer}
\end{tabular}
\end{center}
\end{table}

\subsubsection{Instance Division}
The advantage of the instance division algorithm is that it decreases the complexity of the original instance by separating the original instance into several sub-instances. Since sub-instances have relatively smaller sizes, they converge faster. For ordinary $Q$-learning, this prevents the instance to stop while it hasn't reached the point of convergence. However, the division algorithm assumes that certain parts of the instance should be scheduled together, which may not be the optimal case. One strategy for this problem is to separate the instance by operation time, so that each sub-instance will have a similar expected finish time for each job, hence more likely to be the optimal case. Even using this strategy, the more sub-instances it splits into, the less optimal the final policy will be. We can see from Table~\ref{tab:Convtimes}, with classical $Q$-learning algorithm, we fail to converge for large instances ($Env_l$ and $Env_{xl}$). By applying instance division algorithm, we are able to decrease the convergence time to the level of small instances ($Env_s$ and $Env_m$).
\subsection{Analysis}
To analyze the performance of our RL solution, we first examine the training and testing process. We then compare the key statistics obtained (makespan, CPU Time) from our solution to results from the baseline method implemented. Last, we compare our solution with results retrieved from RL literature.
\subsubsection{Convergence Time}
\begin{figure}[htbp]
\centerline{\includegraphics[width=8cm]{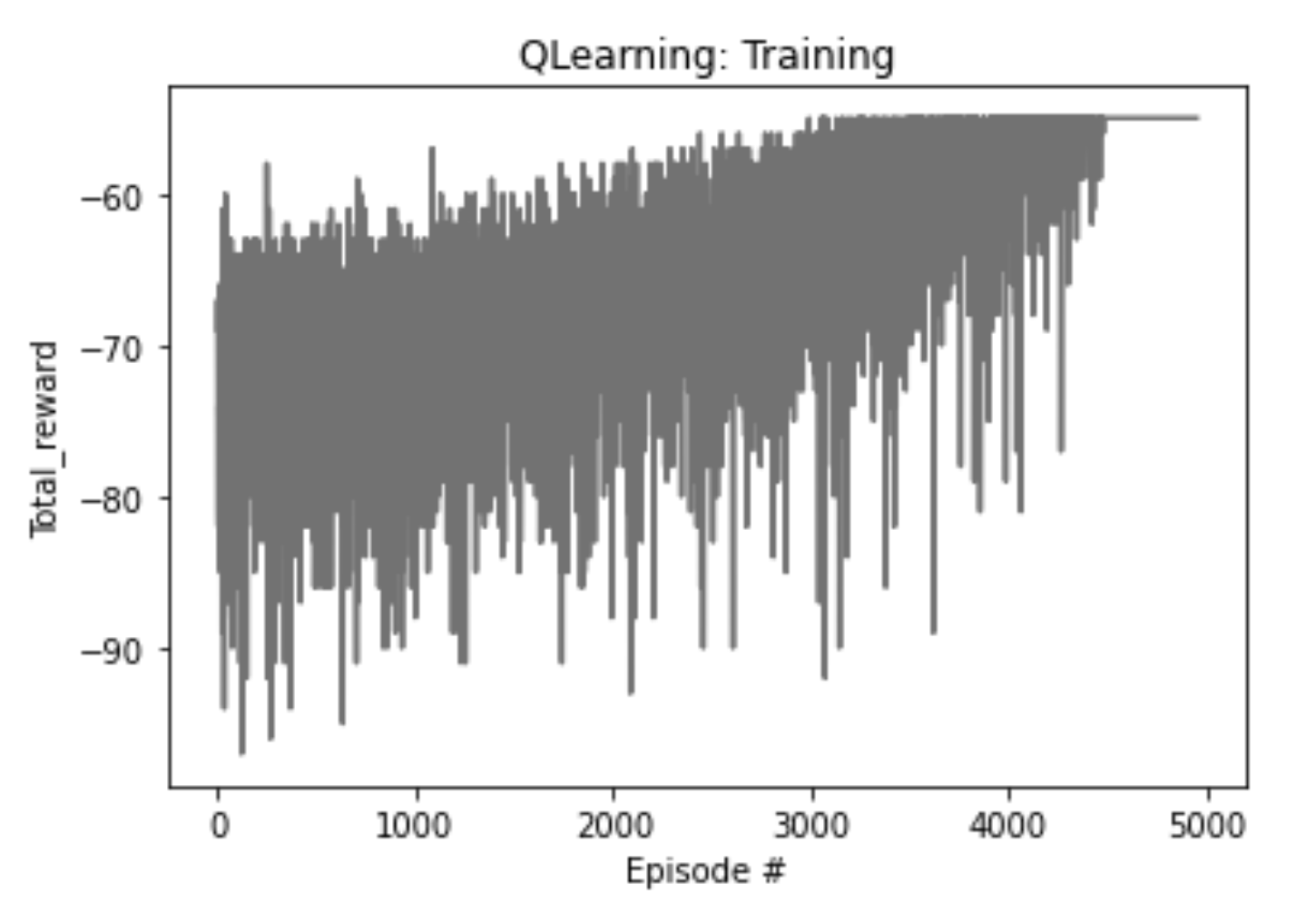}}
\caption{$Q$-Learning Training Process for Medium Environment (6x6)}
\label{fig:training}
\end{figure}

\begin{figure}[htbp]
\centerline{\includegraphics[width=8cm]{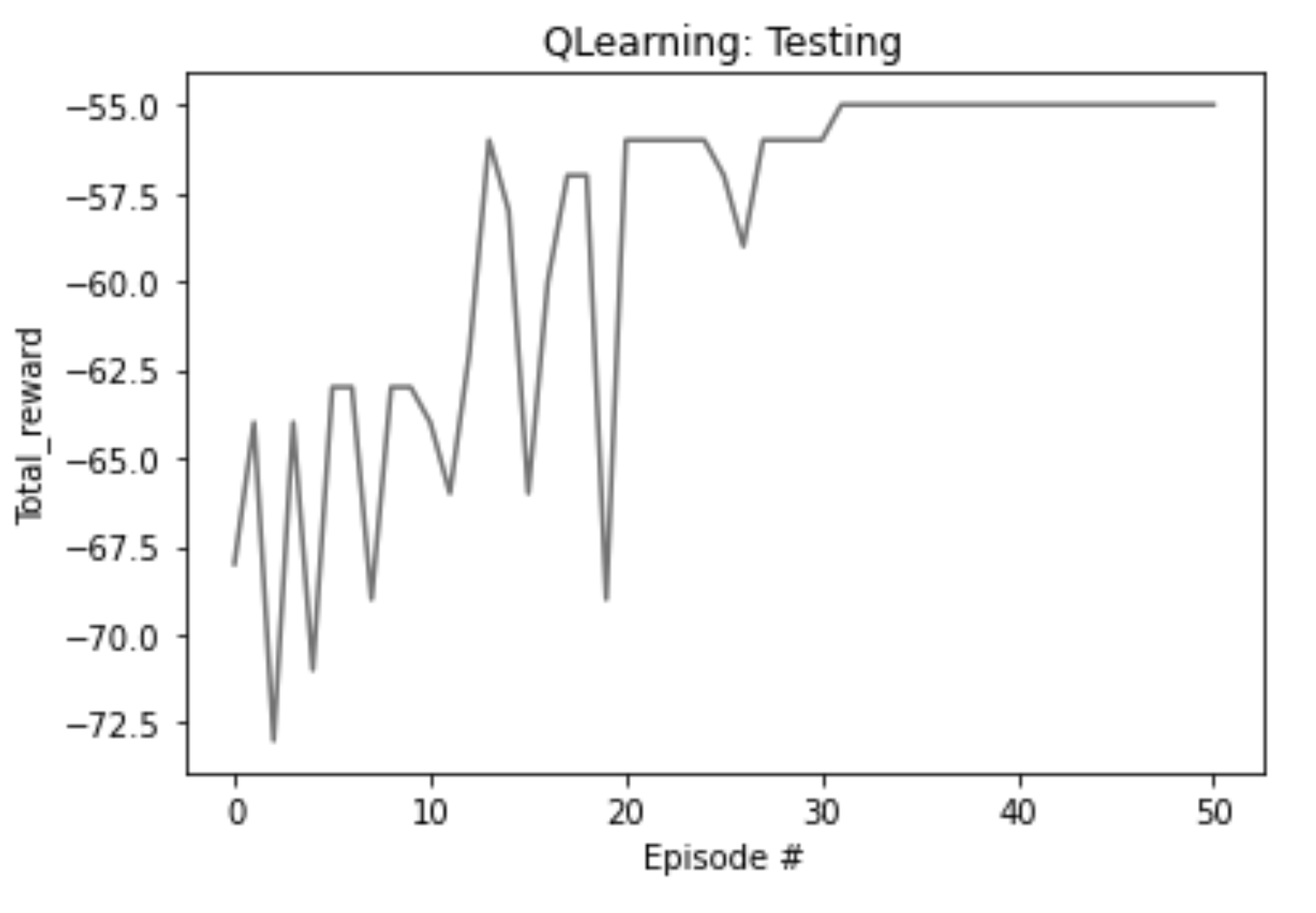}}
\caption{$Q$-Learning Testing Process for Medium Environment (6x6)}
\label{fig:testing}
\end{figure}

Our heuristic-guided RL solution is based on the $Q$-Learning algorithm with epsilon greedy policy. In addition, we implemented an epsilon decay method to encourage convergence toward the end of the training process. As designed, the training process starts with large epsilon and high volatility to encourage exploration at the beginning. However, as a result of the policy in place, the process ends with small epsilon and encourages exploitation(Fig.~\ref{fig:training}). 
In addition to the training process, we implemented a testing process every 100 episodes. The testing process removes all randomness from the algorithm and selects the best schedule from training(Fig.~\ref{fig:testing}). 
\subsubsection{Comparison with baseline methods}
To evaluate the performance of our RL solution, we first used benchmark instances from the OR-Library \cite{beasley_1990}. We use makespan $C_{max}$ to evaluate the solution accuracy and CPU time to evaluate the solution efficiency (RL refers to the heuristic-guided $Q$-Learning solution we implemented). All solutions are running on an M1 Pro chip.

As shown in Table~\ref{tab:makespan}, our solution has better accuracy than random sampling, classical dispatching rules (FIFO and MWKR) and integrated approach (GA) in all environments while achieving similar performance as the state-of-the-art solution (OR-Tools). Furthermore, the results in Table~\ref{tab:CPUtime} show that our solution is more efficient than any non-RL methods but is less efficient than OR-Tools on larger instances due to increasing complexity. 
\begin{table}[htbp]
\caption{Makespan: Baseline Selection}
\label{tab:makespan}
\begin{center}
\begin{tabular}{llllllll}
\br
Env & 
Size & 
RS & 
FIFO & 
MWKR & 
GA & 
RL & 
OR-Tools \\
\mr
$Env_{s}$& 2x3& 53& 58& 61& 53& 53& 53 \\
$Env_{m}$& 6x6& 55& 72& 68& 55& 55& 55 \\
$Env_{l}$& 10x6& 47& 65& 70& 42& 42& 40 \\
$Env_{xl}$& 10x11& 990& 1156& 1096& 986& 933& 927 \\
\br
\end{tabular}
\end{center}
\end{table}
\begin{table}[htbp]
\caption{CPU Time (s): Baseline Selection}
\label{tab:CPUtime}
\begin{center}
\begin{tabular}{lllllll}
\br
Env & 
RS & 
FIFO & 
MWKR & 
GA & 
RL & 
OR-Tools \\
\mr
$Env_{s}$&  0& 0& 0& 0& 0& 0 \\
$Env_{m}$&  7& 0& 0& 12& 5& 0 \\
$Env_{l}$&  132& 4& 4& 833& 127& 0 \\
$Env_{xl}$&  500+& 13& 14& 500+& 183& 3 \\
\br
\multicolumn{7}{l}{All results rounded to the closest integer}
\end{tabular}
\end{center}
\end{table}
\subsubsection{Comparison with RL Solutions from Literature}
To further evaluate our solution, we used instances referenced in other RL literature and compared the makespan obtained from our solution with the results from other literature. The OR-Tool results are used as optimal solutions for reference. The corresponding literature for each instance is mentioned in VI.A. Benchmark Instances.
\begin{table}[htbp]
\caption{Makespan: RL Solution from Literature}
\label{tab:literature}
\begin{center}
\begin{tabular}{lllll}
\br
Env & 
Size &
Literature & 
RL & 
OR-Tools \\
\mr
la01&  10x5& 666& 614& 609 \\
la02&  10x5& 667& 655& 655 \\
la03&  10x5& 597& 590& 588 \\
la04&  10x5& 590& 558& 558 \\
la05&  10x5& 593& 593& 593 \\
Mk04&  15x8& 66& 66& 60 \\
Mk05&  15x4& 173& 155& 127 \\
Ta01&  15x15& 1278& 1266& 1231 \\
Ta41&  30x20& 2208& 2376& 2144 \\
\br
\end{tabular}
\end{center}
\end{table}
As shown in Table~\ref{tab:literature}, our solution outperforms methods from other literature with one exception, Ta41. There are multiple reasons for the performance difference between the method used in \cite{tassel_gebser_schekotihin_2021}(literature which uses Ta41 instance) and our solution. First, \cite{tassel_gebser_schekotihin_2021} focuses on JSSP while our solution is designed for FJSSP. Ta41 is a JSSP instance with one machine per operation which we use as a special case of FJSSP. In addition, \cite{tassel_gebser_schekotihin_2021} presents a Deep Reinforcement Learning(DRL) method which allocates more computational resource. Hence why a difference in performance on Ta41 instance. 

\section{Conclusion and Future Work}
\label{section:conclusion_and_future_work}
In this study, we present a novel OpenAI gym environment for JSSP and FJSSP with efficient search-space reduction. Using this environment, we offer two improvement methods, heuristic guidance and instance division, to the classical Q-learning solution. The results obtained outperformed classical dispatching rules, integrated methods and RL solutions from other literature in both accuracy and efficiency. In addition, our solution has state-of-the-art performance close to the best constraint solver (OR-Tools).

For future works, we aim to integrate stochasticity(such as job release time, machine breakdown, and stochastic duration) into our environment for further study on the robustness and adaptivity of RL solutions. Adding stochasticity also helps with the simulation of practical environments considering the unpredictability of scheduling problems in real life. Furthermore, we plan to improve the efficiency of our solution as it is shown in the evaluation that the OR-Tools solution is significantly more efficient in large instances.

Last, we find it necessary to introduce Deep Reinforcement Learning into our solution to improve the usage of computational resources. The integration of deep learning can help our solution tackle larger problems with extensive complexity. We believe Deep Reinforcement Learning with optimizing techniques is capable of solving JSSP and FJSSP with unprecedented accuracy and efficiency. 

\section*{Acknowledgment}
Hongjian Zhou and Boyang Gu contributed equally to this work and should be considered co-first authors. 
We would like to express our sincere gratitude to Professor Nick Hawes from the University of Oxford for supervising and assisting with our research. 

\bibliographystyle{iopart-num}
\bibliography{Sources}

\end{document}